\documentclass[journal]{IEEEtran}
\usepackage{blindtext,graphicx,booktabs,amsmath,amssymb,bbding,subfig,color}
\IEEEoverridecommandlockouts

\title{A Tempt to Unify Heterogeneous Driving Databases using Traffic Primitives}

\author{Jiacheng Zhu$^{1}$,Wenshuo Wang$^{1}$, Ding Zhao$^{1,*}$

\thanks{This study is supported by Denso International America, Inc. }

\thanks{ $^{1}$J.  Zhu, W. Wang and D. Zhao (\texttt{corresponding author: zhaoding@umich.edu}) are with the Department of Mechanical Engineering, University of Michigan, Ann Arbor, MI 48109, USA. }

}

\begin{document}
\maketitle
\begin{abstract}
A multitude of publicly-available driving datasets and data platforms have been raised for autonomous vehicles (AV). However, the heterogeneities of databases in size, structure and driving context make existing datasets practically ineffective due to a lack of uniform frameworks and searchable indexes. In order to overcome these limitations on existing public datasets, this paper proposes a data unification framework based on traffic primitives with ability to automatically unify and label heterogeneous traffic data. This is achieved by two steps: 1) Carefully arrange raw multidimensional time series driving data into a relational database and then 2) automatically extract labeled and indexed traffic primitives from traffic data through a Bayesian nonparametric learning method. Finally, we evaluate the effectiveness of our developed framework using the collected real vehicle data.
\end{abstract}


\IEEEpeerreviewmaketitle

\section{Introduction}

The last decade has witnessed a dramatic growth in autonomous vehicles (AV) with many learning-based algorithms and techniques in terms of decision-making, path-planning, and optimal control. High-quality naturalistic driving data is required to test and evaluate these approaches \cite{timappenzeller}. Most existing traffic or driving datasets are chronologically logged and organized and thus an extensive post-processing is required for users to extract interested information \cite{zhao2017trafficnet}. However, the heavy, tedious data processing work usually discourages the potential users, hence downgrading the usage of these data collected with huge efforts.
Thus, establishing an accessible and flexible framework capable of automatically arranging, unifying and storing heterogeneous driving data is required. The data heterogeneity mainly appears in three types:
\begin{itemize}
\item Between datasets;
\item Within datasets;
\item Existing datasets and on-going collecting data.
\end{itemize}
\subsubsection{Between Datasets}
The existing naturalistic driving datasets are largely heterogeneous due to diversity in driving settings and data collection platforms. Researchers usually collect data using vehicles equipped with a set of sensors such as camera, Lidar, radar, GPS and IMU, etc. Each sensor records data in specific format and frequency, which will lead to heterogeneity between two datasets, even for data with same labels. There exist many public data benchmarks for different applications and most of them are independently accessible and well organized \cite{yin2017use,wang2017much}; however, these datasets greatly differ from each other in terms of format, traffic context, application, sensor statue, support tools, as shown in Table \ref{Table:parameters}. For example, KITTI Vision Benchmark Suit \cite{kitti_geiger2013vision} is one of the most prestigious datasets for self-driving car applications, including SLAM, object detection and tracking, and road/lane detection. It contains PNG images and TXT files for Velodyne and GPS/IMU data, and XML for bounding box labels. While the Ford Campus Vision and Lidar Dataset \cite{pandey2011ford} lays in diversified sensor configuration, including high precision localization devices and multiple-LiDARs. The raw data is stored in the shape of PPM, LOG, and MAT, which can be directly accessed via C/MatLab. Udacity also provides a high quality dataset as an open source project, with images in PNG/JPEG,  labels in CSV files, and raw data in ROSBAG. 

In addition to the public dataset benchmarks, some  original equipment manufacturers and automobile suppliers provide hardware and proprietary platforms for data collection, for example, BMW's Connected Drive \cite{bmw}, Audi Connect \cite{audi} and GM's OnStar \cite{gm}. The academic researchers have also made some great efforts, for example, MIT's CarTel \cite{hull2006cartel} project, University of Michigan's TrafficNet \cite{zhao2017trafficnet}, SafetyPilot \cite{bezzina2014safety}, Carlab \cite{pese2017carlab} and UMass' DOME \cite{soroush2009dome}. These data-driven applications are all well integrated but lack a standard framework to access data between them.

\begin{figure*}[t]
	\centering
	\includegraphics[width=\linewidth ]{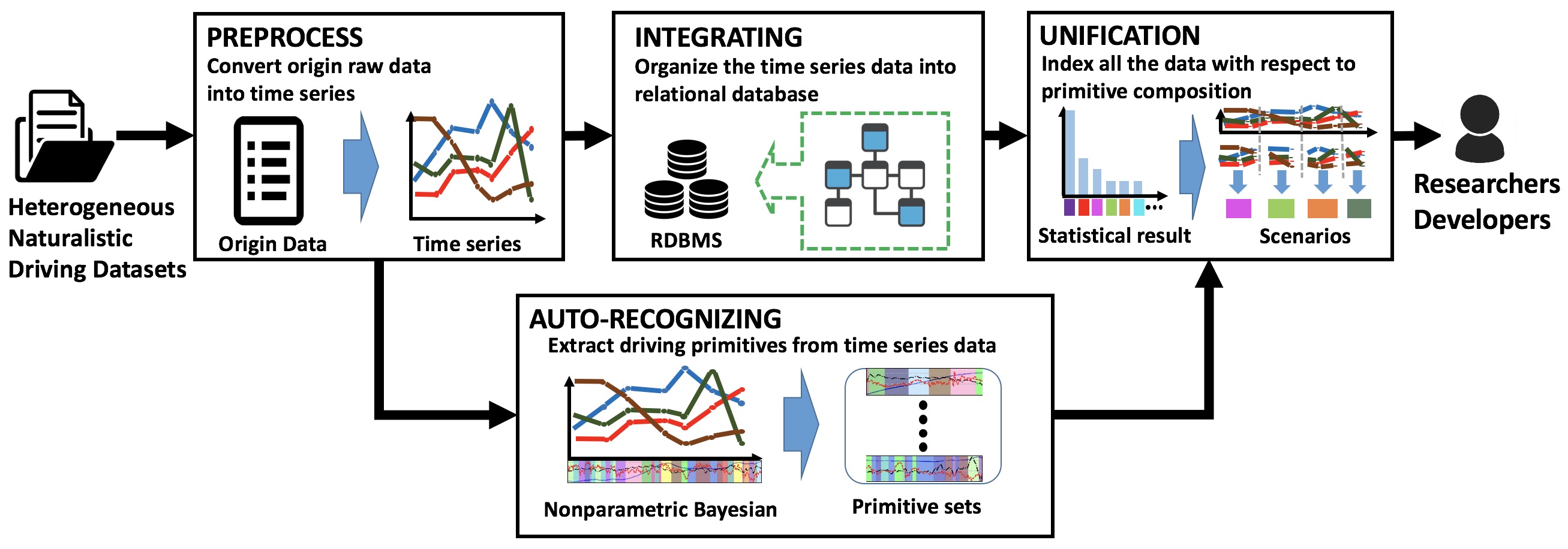}
	\caption{Architecture of auto-recognizing heterogeneous naturalistic driving datasets.}
	\label{fig:framework}
\end{figure*}


\begin{table*}[t]
 \centering
 \caption{\textsc{Comparison of Some Existing Databases}} \label{Table:parameters}
 \begin{tabular}{ccccccc}
  \hline
  \hline
  Name & Format  & Sensor & Scenario Based & Query Oriented & Benchmark & Auto-label\\
  \hline
  KITTI\cite{kitti_geiger2013vision}  & png, txt, xml  & Camera, Lidar, GPS, IMU&  &   & \checkmark & \\\hline
  Ford\cite{pandey2011ford} & mat,ppm,log, pcap & Camera, Lidar, GPS, IMU & & & &    \\	\hline
  Udacity & jpg, log, csv, ROSBAG & Camera, Lidar, GPS,IMU & & & \checkmark & \\ \hline
  Safety Pilot\cite{bezzina2014safety} & csv & Mobieye, GPS, IMU, Radar & & \checkmark &  & \\ \hline
  CarLab\cite{pese2017carlab} & csv, json,SQL & GPS, IMU, driver behavior & \checkmark & \checkmark & &   \\ \hline
  CarTel\cite{hull2006cartel} & SQl, image, video & Camera, GPS, IMU, Chemical & & \checkmark &  \\ \hline 
  TrafficNet\cite{zhao2017trafficnet} & csv, SQL & Mobieye, GPS, IMU, Radar & \checkmark & \checkmark & & \\ \hline
  Our Framework & jpg, csv, SQL, ROSBAG & Camera, Lidar, Radar, GPS, IMU & \checkmark & \checkmark & \checkmark & \checkmark \\
  \hline
  \hline
 \end{tabular}
\end{table*}

\subsubsection{Within Datasets}
In addition to between datasets, the heterogeneity also exists within one single dataset. A variety of traffic scenarios should be contained to investigate and test algorithms under different working conditions, which will cause the diversity in vehicle behaviors and sensor records and hence two parts from this dataset may greatly different from each other.  \par

\subsubsection{Existing datasets and on-going collecting data}
Lastly, the existing datasets could also be greatly different from the on-going collection data due to uncertainty and diversity of driving scenarios. Research demonstrates that the heterogeneity of existing driving datasets will increase greatly at the begin and then increase slightly after more enough data collected in terms of their statistical metrics. \par

A well-unified dataset could facilitate to develop autonomous vehicles. Towards this end, this paper proposes a traffic primitive-based framework. First, information in dataset will be rearranged into time series according to data attributes and timestamps. Then a nonparametric Bayesian learning approach will be applied to extract the primitives by segmenting the multidimensional time series. After classification and learning statistical and physical properties of primitives, the traffic scenarios finally will be defined as the composition of primitives, and thus all the data could be labeled and indexed according to traffic scenarios. \par

Attribute discontinuity and synchronous mistakes of time-series data will make it inflexible to extract primitives from multidimensional time series. Besides small amount of data also makes it difficult to find representative primitives and scenarios. 
In order to make this unification framework tractable, a set of traffic datasets should be eligible in the two following capabilities: 
\begin{enumerate}
\item {\it Recognizability}, which refers to whether the heterogeneity within a single dataset could be eliminated. 
\item {\it Comparability}, which describes the capability of unifying a set of different datasets.  
\end{enumerate}
Tagging and labeling a huge amount of driving data with a variety of aspects and parameters is still a challenging task, which impedes the development and evaluation of autonomous vehicles. For example, 800 human hours are required to manually label data for every one hour recorded data using deep learning techniques \cite{deepai}. Moreover, the manually labeled or extracted data might also not be suitable to learn a model because of large biases. By using nonparametric Bayesian approaches in this paper, we can not only create a unify index for traffic data but also use this index as label. Since the nonparametric Bayesian method segments and classifies the data according to its inherent characteristic\cite{wang2018extracting}, the extracted primitives can represent meaningful real world traffic scenarios.
After all, a well-designed Relational Database Management System (RDBMS)\cite{maier1983_relational_database_theory} is utilized to uniformly arrange and index big data according to defined driving scenarios.
%
%


The remainder of this paper is organized as follows. Section II introduces the database architecture. Section III shows a demonstration of storing and applying unification framework to certain data. Section IV makes further conclusion and discussion.

\section{Data Unification Architecture}


\subsection{Architecture Overview}

The schema of our proposed data unification framework is given in Fig. \ref{fig:framework}, which consists of four modules: preprocessing, integrating, auto-recognizing, and unification. 
\begin{enumerate}
\item Preprocessing: This module aims to arrange raw multidimensional driving data into time series, enabling to store data into a RDBMS. Although the data format and frequency vary among datasets, they have the identical feature of standard timestamps, which endows the preprocessing module to unpack, reorder, convert the dataset files to time series.
\item Integrating: This module aims to store all time-series data into a number of tables, where each table is corresponding to one kind of sensor. The extracted scenarios can also be stored in the database thereafter. The relational structure of RDBMS is utilized by the following indexing process, which makes it flexible for researchers and developers to query and maintain the unified and indexed data.
\item Auto-Recognizing:  This module, as the core of this framework, applies a unsupervised learning approach to divide all the preprocessed time-series data into segmentations, called traffic primitives, automatically. Then the primitives will be clustered and classified according to their statistical similarity. 
\item Unification: This module investigates the statistical properties of the primitive sets and also the real world meaning of certain time series data. The traffic primitives will also be filtered according to duration, distribution and appearance frequency. Subsequently, the traffic scenarios, usually the combination of primitives, could also be defined by primitive indexes. Finally, the whole dataset can be presented by the combination of traffic primitives.    
\end{enumerate}



Under this framework, the dataset can be reorganized according to real-world scenarios. In such way, engineers and researchers could easily query data for testing their algorithms in specific environments. For example, if  researchers tend to evaluate the performance of their LiDAR-camera sensor fusion algorithm in the scenario where the vehicle is making a turn, then the developed data platform enables them to easily query data by selecting Lidar and image data with labels of 'turning', rather than downloading a whole package of raw data and preprocessing data to pick interested LiDAR and camera information. 
Besides, our proposed framework can offer an evolving data ecosystem for autonomous vehicle applications since the data in different formats is well-integrated and indexed by scenarios. While the processed traffic dataset is accumulating, it is more likely for the algorithm to detect and recognize more reasonable scenarios, which can continuously evolve and benefit all developers.

\subsection{Traffic Primitive Extraction}


In order to provide well-labeled multi-dimensional traffic data, traffic primitive that represents principal compositions of driving scenarios should be recognized and extracted \cite{wang2018extracting}. Powerful unsupervised learning-based technologies have been developed to achieve this. For example, Bender, {\it et al}. \cite{bender2015unsupervised_extract_9} applied a Bayesian multivariate linear model to segment a two-dimensional driver behavior data in time-series. Taniguchi, {\it et al}. \cite{taniguchi2015unsupervised_extract_11,taniguchi2016sequence_extract_10} proposed a double articulation analyzer based on nonparametric Bayesian theory for predicting six-dimensional data. Hamada, {\it et al}. \cite{hamada2016modeling_extract_12} raised a nonparametric Bayesian approach with linear systems to learn and predict driver behaviors. Wang, {\it et al}. \cite{wang2018drivingstyle} investigated three different nonparametric Bayesian approaches to analyze drivers' car-following styles. Apart from previous research focusing on driver's behavior, wang, {\it et al}. \cite{wang2018extracting} proposed a framework to extract primitives from multi-scale high-dimension raw traffic data and generate traffic scenarios. In order to automatically extract primitives with less subjective intervention and without prior knowledge then find and cluster the analogous, they introduced nonparametric Bayesian learning based on the sticky HDP-HMM method.

\begin{figure}[t]
	\centering
	\includegraphics[width=0.45\linewidth]{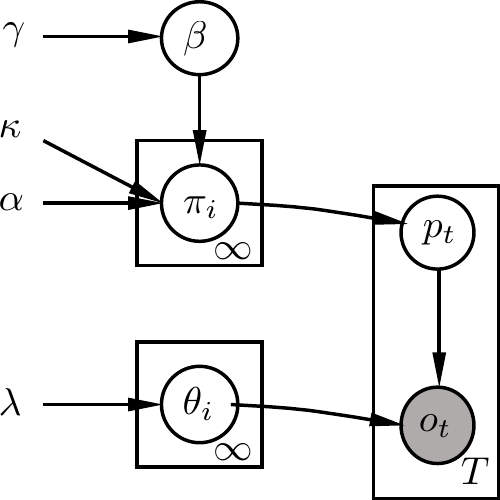}
	\caption{Graphic illustration of the sticky HDP-HMM.}
	\label{fig:extract}
\end{figure}

The entire driving process can be treated as a logic combination of primitives and hence the dynamic process among primitives can be treated as a probabilistic process. Here the traffic were modeled as a dynamic process of primitives based on hidden Markov models (HMM), which consists of two layers: hidden layer and observation layer. The hidden layer represents traffic primitives $p$ and observation layer represents the collected data.

Given time-series observations $\textbf{O}={\{o_{t}\}}^T_{t=1}$ with $\textbf{O} \in \mathbb{R}^{d\times T}$, each traffic primitive $p_{t}$ at time $t$ will be subject to one entry of $\mathcal{O}$. The transition probability from traffic primitives $p_i$ to $p_j$ is denoted as ${\pi}_{i,j}$, where ${\pi}_i = [ {\pi}_{i,1},{\pi}_{i,2},{\pi}_{i,3}...] $. The observation $o_t$ at time $t$ is generated by emission function $o_t=F(o_t | p_t)$. Thus, the HMM can be described as  

\begin{subequations}
\begin{align}
p_t|p_{t-1} \sim {\pi}_{p_{t-1}} \\
o_t|p_t \sim F( {\theta}_{p_t} )
\end{align}
\end{subequations}
where $F(\cdot)$ is the emission function and $ {\theta}_{p_t}$ is the emission parameter. Because the intrinsic dynamic of traffic is changing and open ended, the dimension of the parameter space regarding hidden states in the model is believed to be infinite\cite{wang2018extracting,donoso2014foundations_extract_15}. In order to represent such situation, a prior probability distribution on an infinite-dimensional space is introduced. Here the sticky hierarchical Dirichlet Process with HMM (HDP-HMM)($\gamma, \alpha,H$) is employed\cite{fox2011bayesian_driving_17} by adding an extra parameter $\kappa > 0$, thus we have the sticky HDP-HMM as 

\begin{subequations}
\begin{align}
\beta|\gamma & \sim GEM(\gamma)\\
\pi_i|\alpha,\beta,\kappa & \sim DP(\alpha+\kappa,\frac{\alpha\beta+\kappa\delta_i}{\alpha+\kappa} ),i=1,2,... \\
x_t|x_{t-1} &\sim \pi_{x_{t-1}}, t= 1,2,...,T  \\
y_t|x_t, \theta_{x_t} &\sim F(\theta_{x_t}), t=1,2,...,T \\
\theta_i | H &\sim H, i=1,2,...
\end{align}
\end{subequations}
where $T$ is the data length. In order to make the algorithm tractable for deriving all necessary expression to perform inference\cite{ryden2008versus_extract_18}, we assume that observation are drawn from a Gaussian distribution\cite{mahboubi2017learning_extracting_16,fox2011bayesian_driving_17} with $\theta_i = [\mu_i, \sigma_i]$. As a result, if the priors for observations and transition distributions are learned correctly, the full-conditional posteriors can be learned correctly and the full-conditional posteriors can be computed using Gibbs sampling method.

With the method introduced above, we can parse a long-term multi-dimensional time-series data into finite primitives and simultaneously cluster them without requiring the prior knowledge of traffic primitives.



\begin{figure}[t]
	\centering
    \subfloat[Database entity relation diagram]{\includegraphics[width=0.35\linewidth]{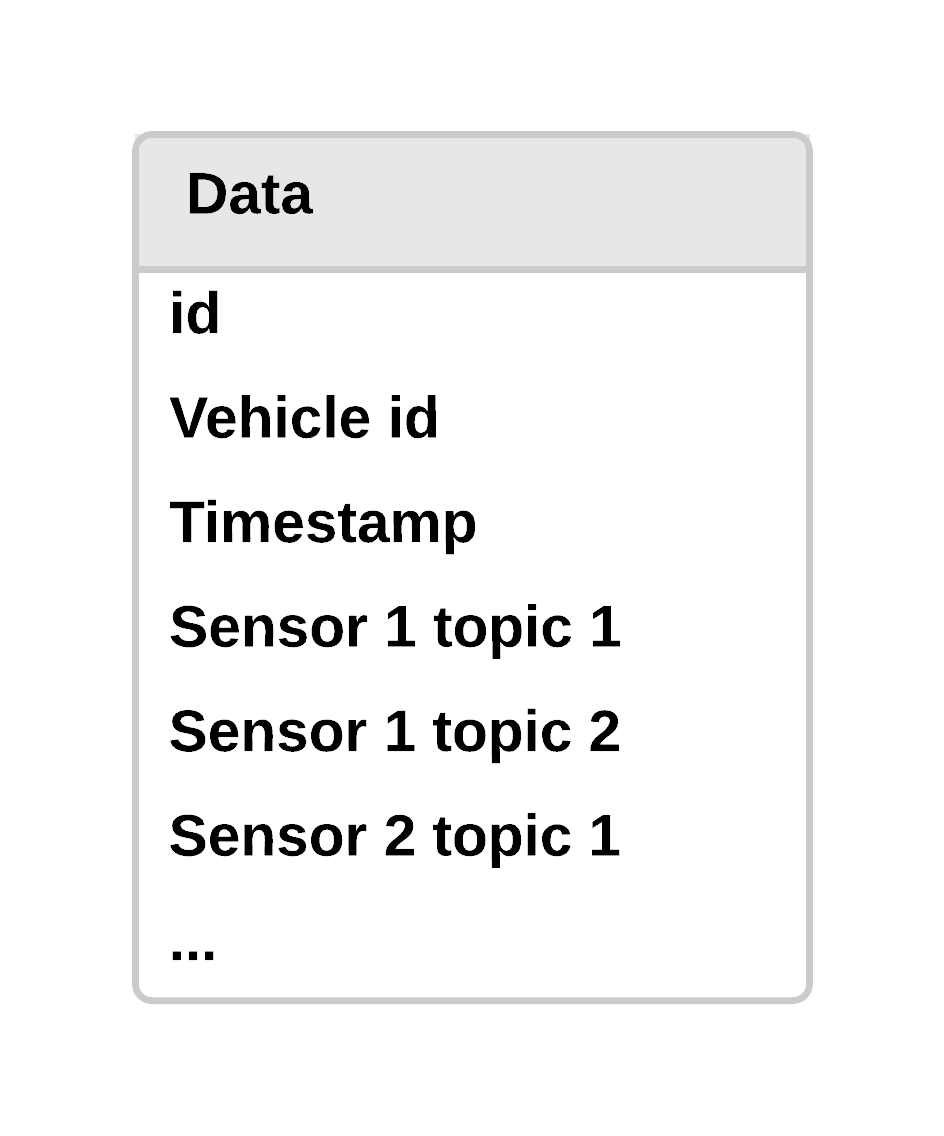}\label{fig:er1}}
    \hfill
	\subfloat[Modified entity relation diagram]{\includegraphics[width=0.65\linewidth]{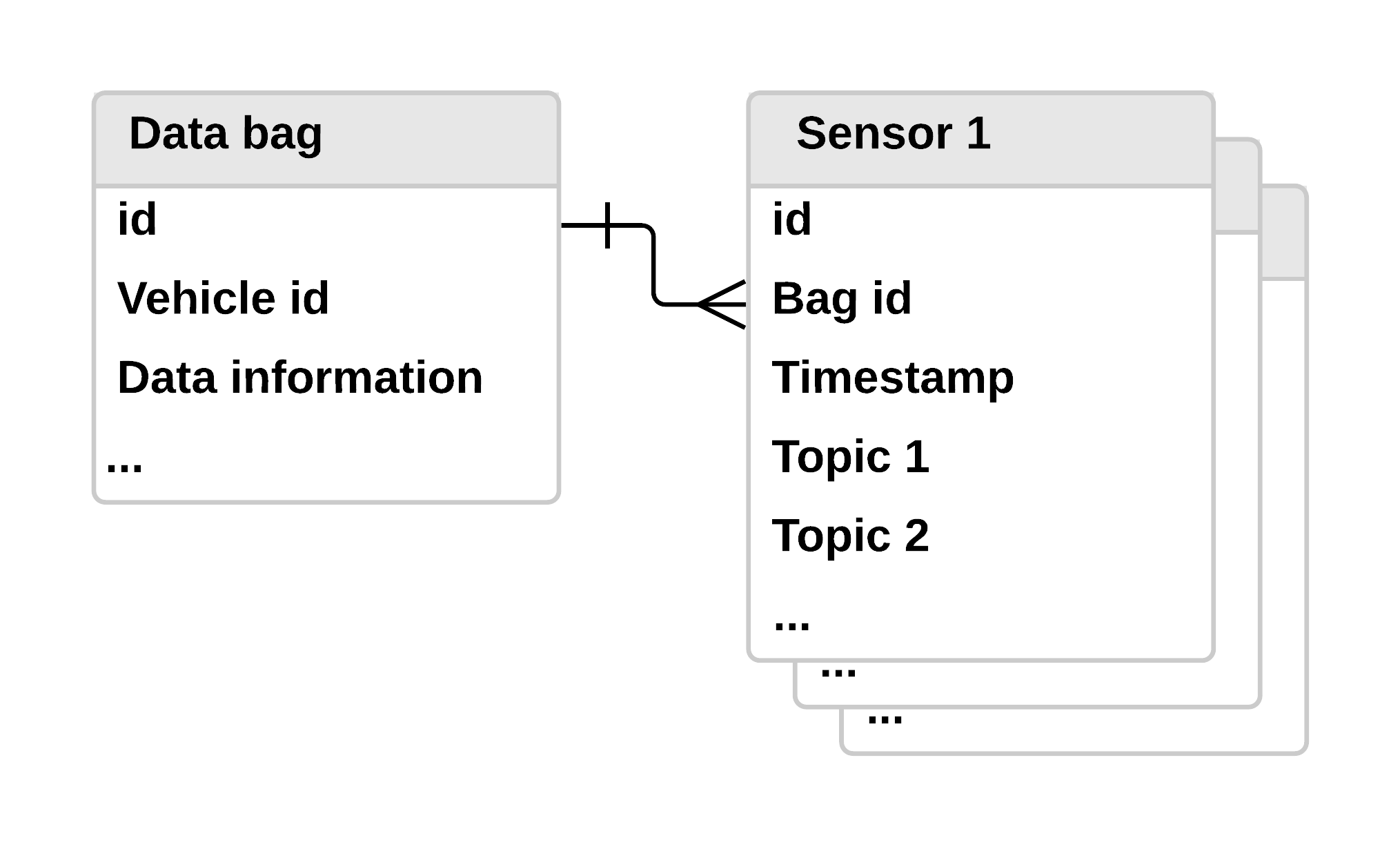}\label{fig:er2}}
	\caption{Current autonomous vehicle database schema.}
	\label{fig:er1ander2}
\end{figure}

\begin{figure}[t]
	\centering
	\includegraphics[width=\linewidth]{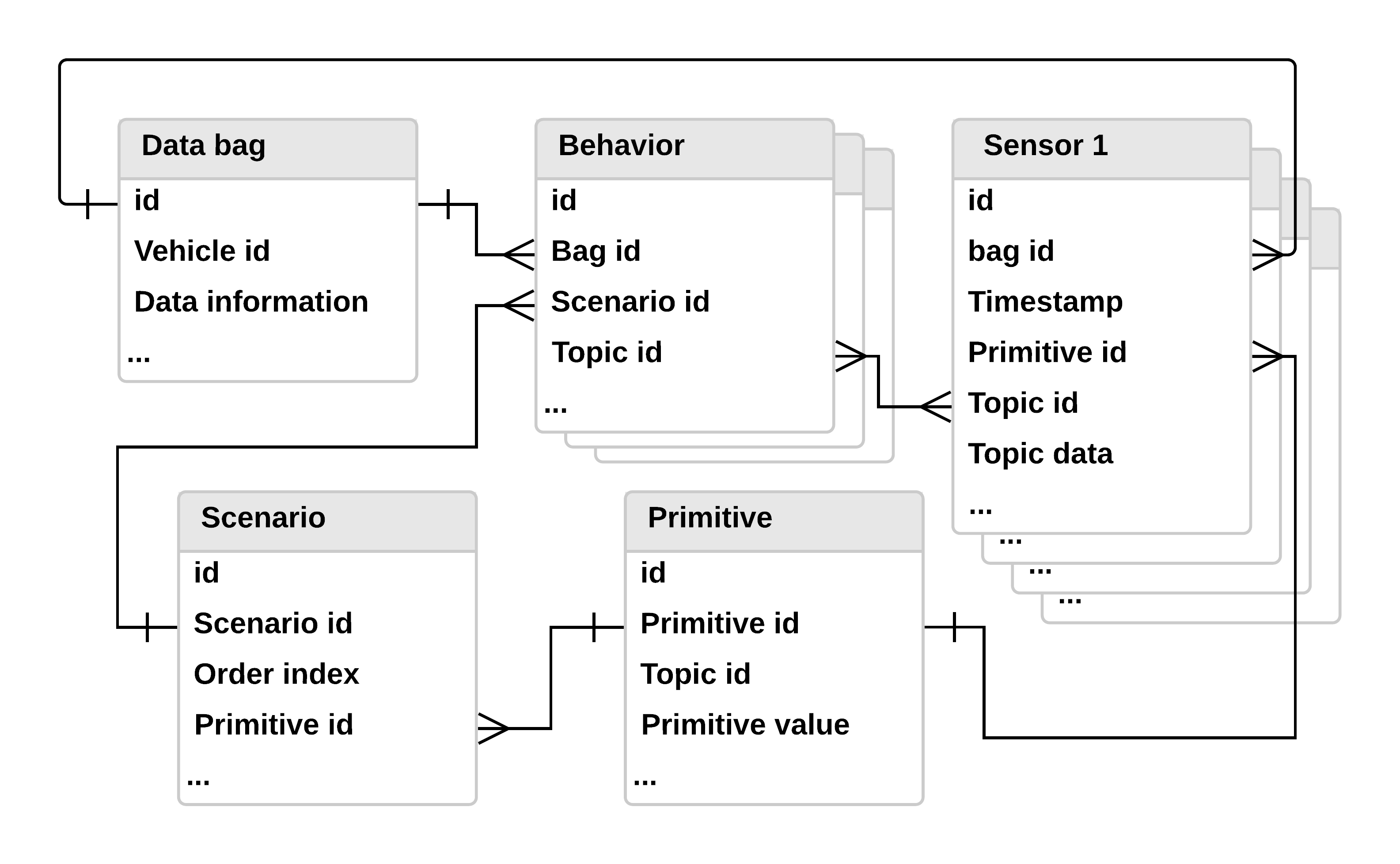}
	\caption{Scenario-based database entity relation (ER) diagram.}
	\label{fig:er3}
\end{figure}

\subsection{Database Schema}

Our developed framework uses the RDBMS to store and organize large time-series data. In order to organize heterogeneous data, we developed an entity relationship (ER) diagram with consideration of driving scenarios. Multiple sensors (e.g., camera, Lidar, radar, steering wheel, brake, GPS and IMU) with their specific `attributes' are employed to test and evaluate autonomous vehicles. For example, the steering wheel sensor contains `attribute' of actual angle, command angle and steering wheel torque. Therefore, naturalistic driving data could contain a large number of `attribute'. We use the schema in Fig. \ref{fig:er1} to illustrate such a database, where each row represents an attribute in the entity/table, which is simple and practicable. What we need to mention is that the table structure has to change if a new column must be added when new attributes appear, for example, a new sensor is triggered on. However, this will cause two problematic tasks:

\begin{itemize}
\item Large number of empty sets: When uploading data, if a row of data lacks certain attribute, the corresponding position will be filled with an empty value, thus arranging all data in one entity/table, which will leads to lots of empty values.
\item Frequently updating requirement: Frequently updating database structure should also be avoid as it may lead to incompatibility of different versions of same database.
\end{itemize}

In order to overcome these problems, we proposed a one-to-many relationship, which replaces multiple columns in a table with a single column of indexes referencing a new table of all related attributes. Treating all the raw data files as bags and each bag contains time-series raw naturalistic data from different sensors. As shown in Fig. \ref{fig:er2}, each sensor is regarded as an entity and its recorded data will be stored in corresponding table separately. Under this schema, the sensor table will be updated if the associated sensor is included, which allows one to query interested sensor data from specific datasets, thereby making it more convenient than directly downloading the whole dataset.
%
%

In order to relate all the data to driving scenarios, we added `Scenario' table, `Primitive' table and `Behavior' table into the developed schema, as shown in Fig. \ref{fig:er3}.
%
%

Here, we do not need prior knowledge of driving scenarios and primitives, but we do need  manually define several hypothetical scenario conditions. 
For example, whether a vehicle is turning left/right or going straight only depends on the vehicle dynamic state and the vehicle control input with regardless of perception behavior. 
Based on this idea, scenarios are categorized into several `Behavior' tables\footnote[1]{It defines the required data topics or dimension for Bayesian nonparametric unsupervised learning.}, which stores scenarios extracted from all datasets. Different `Behavior' is carefully defined by combination of sensor. Each scenario is composed by driving primitives.


\begin{figure}[t]
	\centering
	\includegraphics[width=\linewidth]{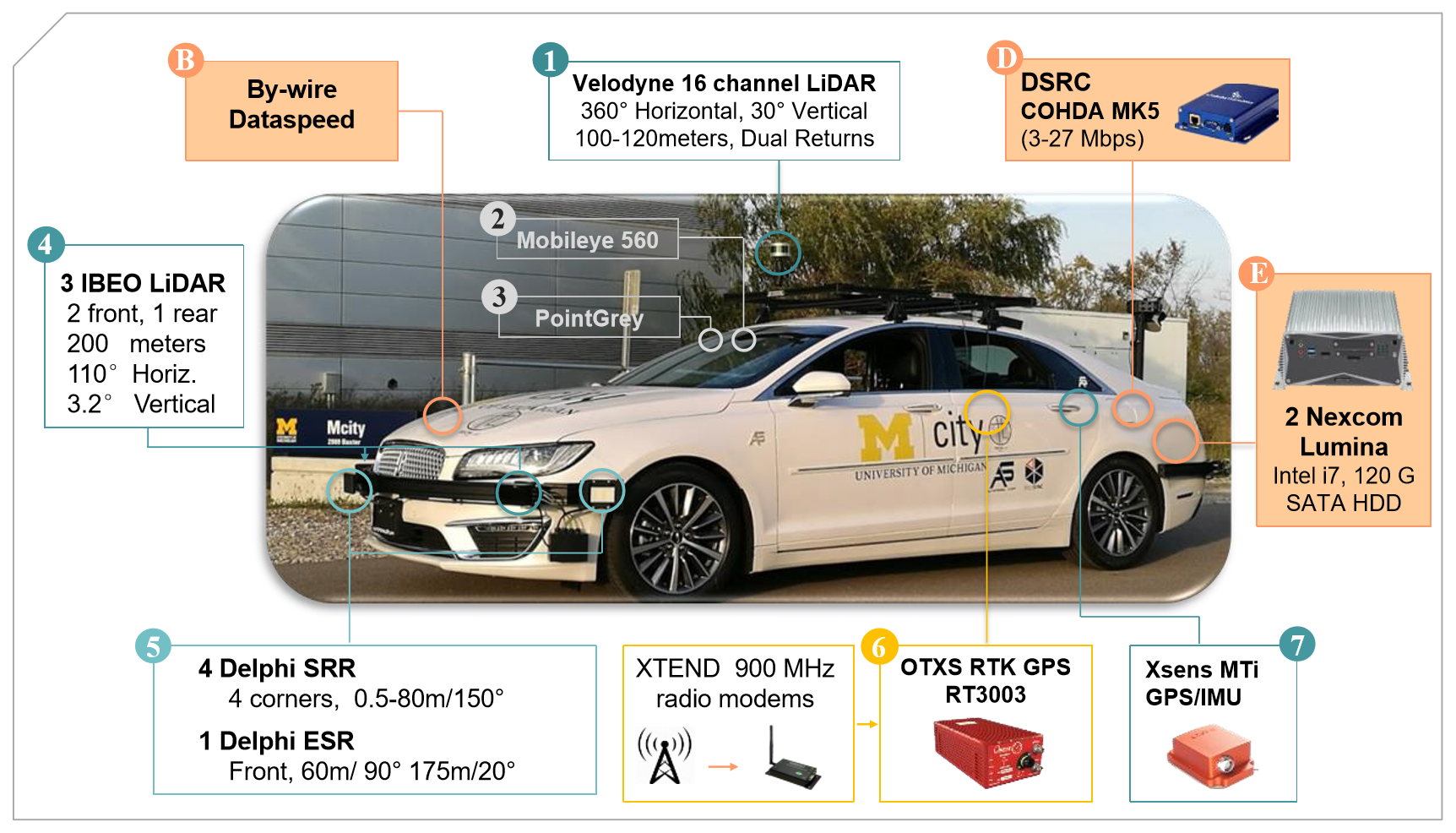}
	\caption{MKZ platform with the equipped sensors and devices.}
	\label{fig:car and sensor}
\end{figure}

\section{Experiment Validation}

The following part will show how our proposed framework categorizes and auto-labels raw data and scenarios. At current stage, our database contains data collected from the University of Michigan open connected and automated research vehicle (\textbf{OpenCAV}) platform.



\subsection{Experiment Equipment}

The costumed Lincoln MKZ CAV research platform is shown in Fig. \ref{fig:car and sensor}, which was equipped with multiple types of sensors, including DSRC, By-wire Control, 2 ECUs, and other ancillary equipment. The operating system of the two ECUs was Linux, Ubuntu 16.04. The Robot Operation System(ROS) was used as the middleware interface. The USB/CAN data and other outfitted sensors data were converted for use with ROS. Each sensor parameter was related to a specific topic and continuously publishes data with timestamps when the platform was running. Once the data collecting process was triggered on, all the published data were recorded. After the collecting process was terminated, the recorded data was packaged into a ROSBAG. 



   
    
    
   
    
   

\begin{figure}[t]
	\centering
	\includegraphics[width=\linewidth]{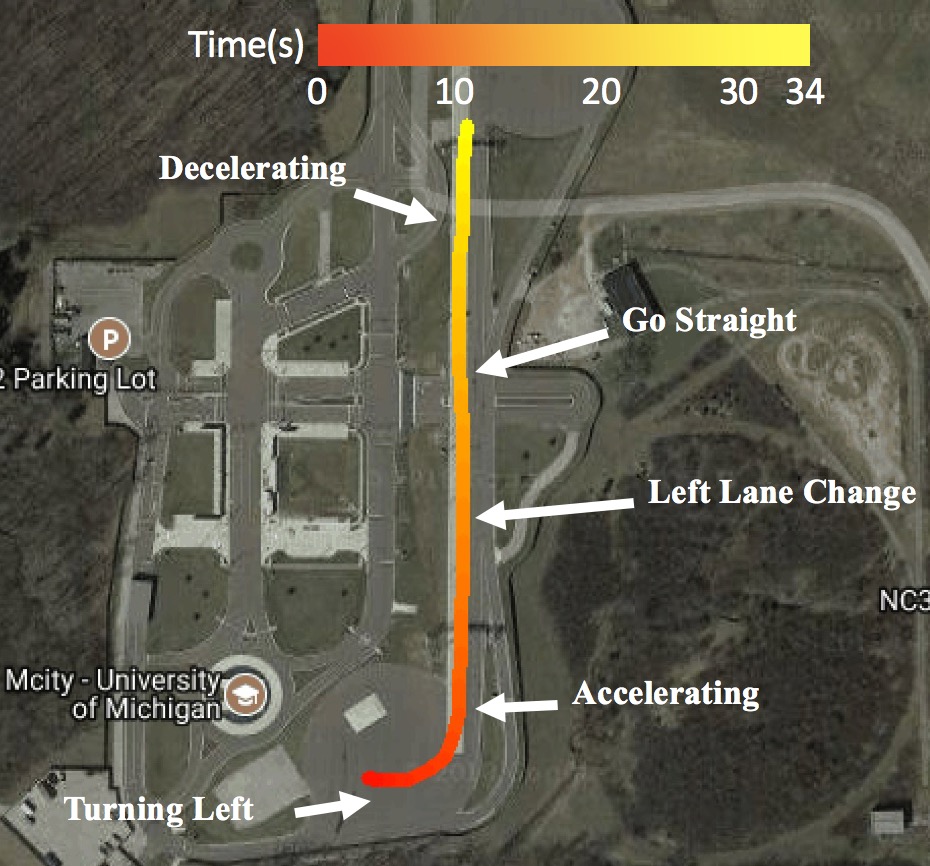}
	\caption{The test trip vehicle trajectory and including scenarios.}
	\label{fig:trajectory}
\end{figure}

\subsection{Testing Situation}

The data collection was performed in the Mcity of University of Michigan. Mcity is a test facility that simulates the broad range of complexities vehicles encounter in urban and suburban environments. The grounds include approximately five lane-miles of roads with intersections, traffic signs and signals, sidewalks, simulated buildings, street lights, and obstacles. \par

Two vehicles, the Lincoln MKZ vehicle and a target vehicle, participated together in order to simulate the naturalistic traffic environment and different driving behaviors. The driver of the target car was asked to drive in certain style and trajectory, then the Lincoln MKZ performed some driving behaviors such as accelerating, car following, lane changing, overtaking, etc.\par

In order to evaluate our data unification framework on extracting meaningful driving primitives, a multiple of different driving scenarios and behaviors were included in a single ROSBAG. All the sensors were tested in good circumstances so little flaw and discontinuity exited, thus guaranteeing the {\it recognizability} of recorded data. The data collection were conducted in peaceful weather, only ordinary traffic scenarios and driving behaviors were performed. Each ROSBAG contains data with duration of no longer than one minute.\par

\subsection{Data Processing and Unification Procedure}

This part will show the unification procedure of one ROSBAG which contains a trip of 34.405 seconds with 1180 frames, as shown in Fig. \ref{fig:trajectory}. During the trip, the Lincoln MKZ conducted the following behaviors:
\begin{enumerate}
\item started and made a left turn;
\item went straight while following the target car;
\item accelerated and changed left;
\item kept the speed, went straight, and overpassed the target car;
\item slowed down and made a right turn.
\end{enumerate}

The data information in ROSBAG is extracted and stored as multidimensional time series via $rospy$\cite{rospy} python package. Specifically, the data from different sensor topics is read as $message$ objects and each $message$ consists of a timestamp and some attributes. A python algorithm was used to convert all the data into time series according to their timestamps and attributes. 

In order to verify the unsupervised segmentation result, a four dimension time series with $steering\_wheel\_angle$, $steering\_wheel\_angle\_cmd$, $steering\_wheel\_torque$ and $speed$ were used to extract primitives. \par

\begin{figure}[t]
	\centering
	\includegraphics[width=1.1\linewidth]{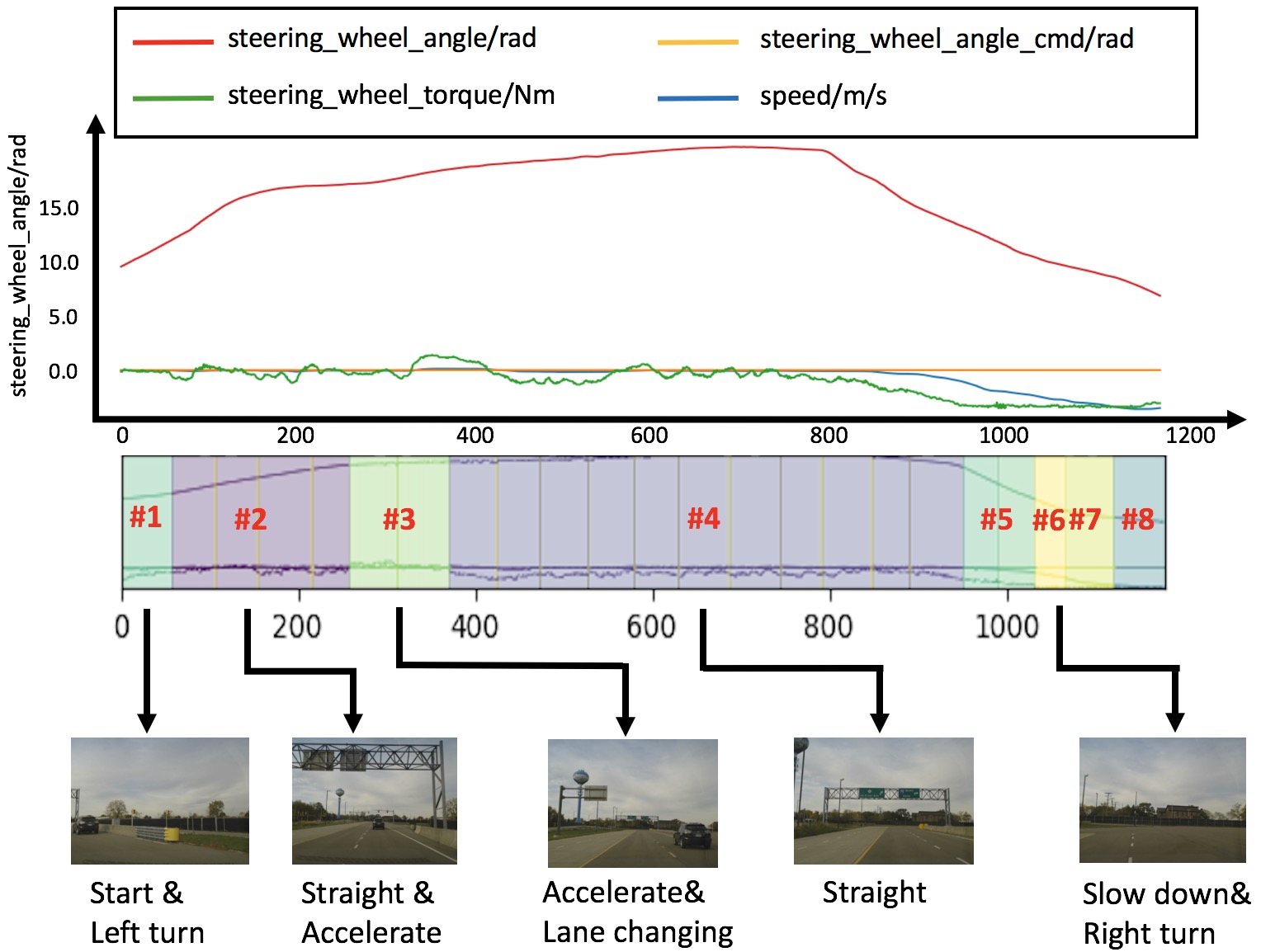}
	\caption{Result of the nonparametric Bayesian learning cluster}
	\label{fig:real_data}
\end{figure}

Fig.\ref{fig:real_data} illustrates that the whole behavior is segmented to eight primitives with different colors. After comparing segmentation result with the steering data and investigating the recorded video, it can be found that primitive \#1 fits the initial low speed left turning, primitive \#2 fits the straight accelerating part, primitive \#3 match the beginning of accelerating and left turning, whereas primitive \#4 satisfies the Lincoln MKZ's long period going straight. The slowing down and turning right can be described by the combination of primitives \#5-\#8. After all, the experimental results demonstrate that the nonparametric Bayesian learning can segment and classified the whole driving behavior reasonably without any prior knowledge of driving primitives.  


After obtaining the primitives of driving behavior, the multi-dimensional vector mean and variance of each primitives were stored in the $primitive$ table. Five $Acting$ scenarios, in our case, as 'Starting\&Turning Left', 'Accelerating', 'Lane Changing', 'Going Straight', 'Slowing\&Turning Right' then were represented by extracted primitives, respectively. The detected scenarios can be compared with the current scenarios stored in the $Acting$ table and then new scenarios can be stored.\par

\section{Conclusion and Future Work}
This paper presents a data platform capable of auto-recognizing, unifying, indexing heterogeneous traffic data for self-driving applications based on traffic primitives. We also designed a scenario-based entity relational database for multidimensional traffic data. A real world experiment was conducted to demonstrate that our proposed traffic primitive-based method is able to transfer road naturalistic driving data into distinguishable and recognizable scenarios. The utility of nonparametric Bayesian unsupervised learning methods to extracted traffic primitive was verified.

This paper primarily demonstrates the effectiveness of our proposed framework of data unification based on traffic primitives. By using our proposed framework, other kinds of unification process such as between databases as well as between the existing database and the on-going collecting data will be conducted in future work. We will also consummate our method and database to include different types of autonomous vehicle data and integrate existing autonomous vehicle dataset.



\section*{Acknowledgment}

The authors would like to thank staff of DENSO R\&D Lab, Ann Arbor. This study is supported by Denso International America, Inc. under master alliance program with University of Michigan.


\ifCLASSOPTIONcaptionsoff
  \newpage
\fi

\bibliographystyle{IEEEtran}
\bibliography{ref.bib}

\end{document}